\documentclass[sigconf]{acmart}
\AtBeginDocument{%
  }

\copyrightyear{2023}
\acmYear{2023}
\setcopyright{acmlicensed}\acmConference[MM '23]{Proceedings of the 31st ACM International Conference on Multimedia}{October 29-November 3, 2023}{Ottawa, ON, Canada}
\acmBooktitle{Proceedings of the 31st ACM International Conference on Multimedia (MM '23), October 29-November 3, 2023, Ottawa, ON, Canada}
\acmPrice{15.00}
\acmDOI{10.1145/3581783.3611734}
\acmISBN{979-8-4007-0108-5/23/10}

\acmSubmissionID{189}

\usepackage{booktabs}
\usepackage{multirow}
\usepackage{tikz-cd}
\usepackage{hyperref}
\usepackage{tikz}
\usepackage[safe]{tipa}
\usetikzlibrary{decorations.markings}
\begin{document}

\title{SelfTalk: A Self-Supervised Commutative Training Diagram to Comprehend 3D Talking Faces}

\author{Ziqiao Peng}
\authornote{Equal contribution.}
\affiliation{\institution{Renmin University of China}
\country{}}
\email{pengziqiao@ruc.edu.cn}

\author{Yihao Luo}
\authornotemark[1]
\affiliation{\institution{Imperial College London}
\country{}}
\email{y.luo23@imperial.ac.uk}

\author{Yue Shi}
\affiliation{\institution{Psyche AI Inc.}
\country{}}
\email{shiyue395171@gmail.com}

\author{Hao Xu}
\affiliation{\institution{Hong Kong University of Science and Technology}
\country{}}
\email{hxubl@connect.ust.hk}

\author{Xiangyu Zhu}
\affiliation{\institution{Chinese Academy of Sciences}
\country{}}
\email{xiangyu.zhu@nlpr.ia.ac.cn}

\author{Jun He}
\affiliation{\institution{Renmin University of China}
\country{}}
\email{hejun@ruc.edu.cn}

\author{Hongyan Liu}
\authornote{Corresponding author.}
\affiliation{\institution{Tsinghua University}
\country{}}
\email{liuhy@sem.tsinghua.edu.cn}

\author{Zhaoxin Fan}
\authornotemark[2]
\affiliation{\institution{Renmin University of China}
\country{}}
\email{fanzhaoxin@ruc.edu.cn}
\renewcommand{\shortauthors}{Ziqiao Peng et al.}


\begin{abstract}
Speech-driven 3D face animation technique, extending its applications to various multimedia fields.
Previous research has generated promising realistic lip movements and facial expressions from audio signals.
However, traditional regression models solely driven by data face several essential problems, such as difficulties in accessing precise labels and domain gaps between different modalities, leading to unsatisfactory results lacking precision and coherence.
To enhance the visual accuracy of generated lip movement while reducing the dependence on labeled data, we propose a novel framework SelfTalk, by involving self-supervision in a cross-modals network system to learn 3D talking faces.
The framework constructs a network system consisting of three modules: facial animator, speech recognizer, and lip-reading interpreter. 
The core of SelfTalk is a commutative training diagram that facilitates compatible features exchange among audio, text, and lip shape, enabling our models to learn the intricate connection between these factors. The proposed framework leverages the knowledge learned from the lip-reading interpreter to generate more plausible lip shapes.
Extensive experiments and user studies demonstrate that our proposed approach achieves state-of-the-art performance both qualitatively and quantitatively.
We recommend watching the supplementary video\footnote{\url{https://ziqiaopeng.github.io/selftalk}}.
\end{abstract}






\maketitle

\section{Introduction}
\begin{figure}[h]
  \vspace{-10pt}
  \centerline{\includegraphics[width=1\linewidth]{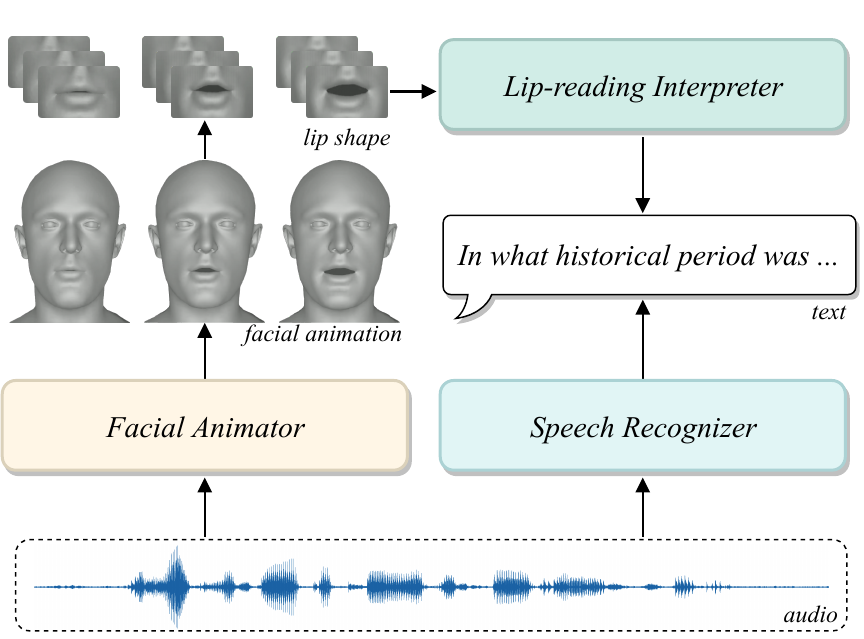}}
  \caption{Given a speech signal as input, our framework can generate realistic 3D talking faces showing comprehensibility by recovering coherent textual information through the lip-reading interpreter and the speech recognizer.}
  \label{fig:1}
  \vspace{-5pt}
\end{figure}
Virtual characters and intelligent avatars are increasingly considered essential components of multimedia applications in various industries such as business, entertainment, and education~\cite{tanaka2022acceptability}.
In this field, Speech-driven 3D facial animation technology~\cite{karras2017audio,pham2017speech} is widely adopted in voice assistants and voice interaction systems to enhance the naturalness and vividness of remote communications.
Therefore, increasing the accuracy and efficiency of 3D talking faces has become a popular research topic in graphics~\cite{lu2021live,cao2022authentic,zhou2018visemenet}, computer vision~\cite{lahiri2021lipsync3d,guo2021ad,fan2022faceformer}, and multimedia technology~\cite{wu2021imitating,hu2021text}.
Previous studies have proposed numerous methods to generate speech-driven 3D face animation. VOCA~\cite{cudeiro2019capture} presents a time-based convolution approach to regress the face movement from audio, the transformer-based autoregressive method used in FaceFormer~\cite{fan2022faceformer} to accommodate considerable audio information.
However, these data-driven regression methods face several essential problems, such as difficulties accessing precise labels and domain gaps between different modalities, leading to unsatisfactory results. Moreover, these methods did not consider the comprehensibility of generated lip movements, which is crucial for human visual perception. \par
%
%
%
To enhance the visual accuracy of generated lip movement while reducing the dependence on labeled data, we propose a novel framework called SelfTalk (Fig. \ref{fig:1}), which involves self-supervision elements in a cross-modal network to generate coherent and visually comprehensible lip movements aligned with natural speech.
Our framework comprises three essential modules, facial animator, speech recognizer, and lip-reading interpreter.
The training processes of three modules form a commutative diagram that facilitates compatible features exchange among audio, text, and lip shape. The commutative diagram~\cite{dwyer1989homotopy} originally represented a graphical representation of mathematical objects and their relationships, where the composition of paths connecting the objects yields the same result regardless of the path taken. This idea enables our modules to learn the intricate connection between these factors in cross-modalities, where the learned information from the lip-reading module will be introduced to generate more plausible lip shapes.\par
The facial animator is the primary step for talking face generation, receiving speech audio as input, and generating 3D face animations as sequences of mesh vertex offsets.
%
The speech recognizer module translates the speech audio into the corresponding text contents. Previous methods have provided several reliable pre-trained models for extracting accurate text from audio. We involve the network proposed by ~\cite{conneau2020unsupervised} and keep its parameters frozen during the entire commutative training diagram. 
The aim of the lip-reading interpreter module is to understand the text content from the lip movement, which expects a coherent result with its corresponding the speech recognizer. 
Therefore, we establish a self-supervised mechanism during training in the above three modules in our framework.
They can collaborate in this framework to generate more precise and visually comprehensible facial movements.

In summary, the main contributions of our work are as follows:
\begin{itemize}
  \item A novel framework called SelfTalk utilizes a cross-modal network system to generate coherent and visually comprehensible 3D talking faces by reducing the domain gap between different modalities.
  \item A commutative training diagram proposed to enhance information exchange among modules and realize the self-supervision of the entire model, which reduces the dependence on labeled data while enhancing the visual accuracy of generated lip movements.
  \item An audio-driven network model to generate precise 3D facial animation with the comprehensibility of lip-reading, presenting state-of-the-art performance.
\end{itemize}

\section{Related Work}
\begin{figure*}[h]
  \centering
  \includegraphics[width=\textwidth]{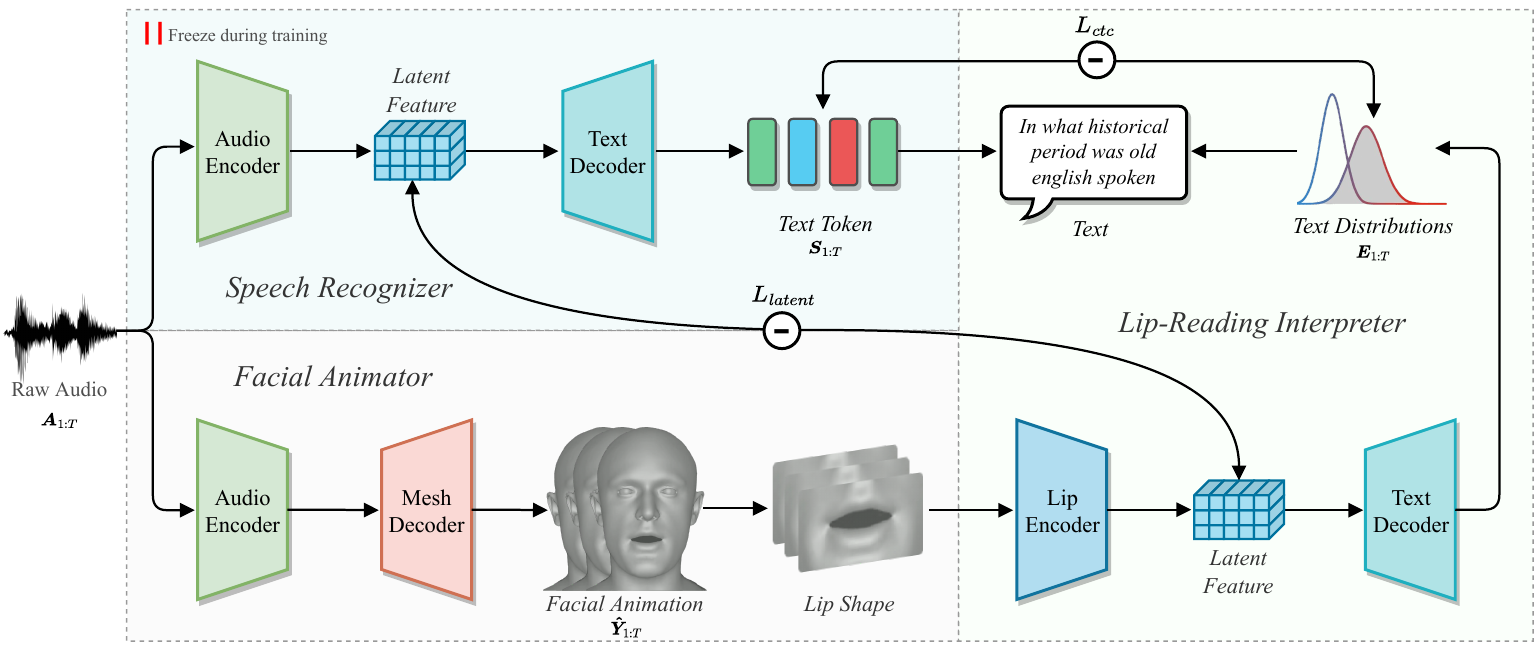}
  \caption{Overview of the proposed SelfTalk. We constructed a commutative training diagram consisting of three modules: facial animator, speech recognizer, and lip-reading interpreter. Specifically, given an input audio signal $\boldsymbol{A}$, the facial animator module extracts the corresponding facial animation $\hat{\boldsymbol{Y}}$, which constitutes the core component of our framework. The speech recognizer, on the other hand, is capable of producing the corresponding text $\boldsymbol{S}$ and utilizing it as a ground truth label for supervision. Lastly, the lip-reading interpreter interprets lip movements, produces a text distribution and establishes a constraint using the label from $\boldsymbol{E}$.}
  \Description{SelfTalk}
  \label{fig:2}
\end{figure*}

\subsection{Speech-Driven 3D Facial Animation}
Speech-driven 3D face animation~\cite{karras2017audio,zhang2010facial,pham2017speech,taylor2017deep,cudeiro2019capture,lahiri2021lipsync3d,richard2021meshtalk,fan2022faceformer,xing2023codetalker,peng2023emotalk} is a field of research that aims to generate realistic facial animation from speech signals. In recent years, many researchers have focused on 2D-based talking face~\cite{wang2023seeing,ji2022eamm,shen2022learning,zhang2022text2video,park2022synctalkface,zhao2022thin,zhang2021facial,guo2021ad,zhou2021pose,wang2021audio2head,zhou2020makelttalk,prajwal2020lip,kr2019towards,siarohin2019first}. Due to the extensive and easily accessible 2D audio-visual dataset, 2D talking face has yielded satisfactory outcomes. In this work, we mainly focused on the driving of 3D talking faces, which can be broadly categorized into view-based and learning-based approaches.

\noindent\textbf{Viseme based 3D face animation.} In the early methods, the view sequence was obtained from input text~\cite{ezzat1998miketalk,ezzat2000visual} or extracted directly from speech using HMM-based acoustic models~\cite{verma2003using}. These methods have advantages for animation control and can be integrated easily into pipelines that are friendly to animators. Visual synthesis is achieved through a blend of templates or context-related view models. For example, Taylor et al.~\cite{taylor2012dynamic} proposed a dynamic view model that uses a one-to-one mapping between phonemes and lip movements. 
JALI~\cite{edwards2016jali} incorporates mouth movements into lip and jaw rig animation to produce compelling co-articulation results. The most recent approach is a parameter view matching algorithm proposed by Bao et al.~\cite{bao2023learning}, which extracts view parameters from speech videos using prior phonetic knowledge. 
These view-based approaches have stronger comprehensibility, but the generated lip shape results are only sometimes accurate and usually require manual adjustments by animators.

\noindent\textbf{Learning based 3D face animation.} The deep learning-based approach relies on large amounts of data and achieves strong lip synchronization results by inputting large datasets during model training. 
Karras et al.~\cite{karras2017audio} proposed an end-to-end CNN to learn the mapping from waveforms to 3D vertices of a facial model. Simulating the talking style of a single actor only requires high-quality animation data of 3-5 minutes. Emotalk~\cite{peng2023emotalk} leverages an emotion-decoupling module to separate speech content and emotion, producing high-quality emotional facial blendshape coefficients.

Moreover, we review the most relevant studies to ours, which involve taking audio as input and producing a 3D mesh. VOCA~\cite{cudeiro2019capture} leverages temporal convolutions and control parameters to produce realistic character animation from speech signals and a static character mesh. 
FaceFormer~\cite{fan2022faceformer} considers the continuity of lengthy sequences and proposes using a Transformer-based model to obtain context-related audio information and generate continuous facial movements autoregressively. CodeTalker~\cite{xing2023codetalker} introduces discrete motion priors by training a vector quantized autoencoder (VQ-VAE)~\cite{van2017neural} to self-reconstruct real facial movements, reducing the problem of over-smoothing facial movements. In contrast, MeshTalk~\cite{richard2021meshtalk} generates a categorical latent space for facial animation that separates audio-correlated and audio-uncorrelated movements through cross-modalities loss. 
However, these methods only focus on training the mapping from speech to face and neglect the intelligibility of lip movements, which can limit the accuracy of lip movements and cause ambiguity for identical pronunciations. In contrast, our method introduces a self-supervised lip reading constraint, which can effectively alleviate this problem and improve the accuracy of lip movements.

\subsection{Lip Reading}
Lip reading is a research area of active interest across multiple fields, including speech recognition, computer vision, and human-computer interaction. In early studies of lip reading, researchers depended on manually crafted pipelines and statistical models to extract visual features and model time~\cite{gowdy2004dbn,livescu2007articulatory,ong2011learning,papandreou2009adaptive}. In recent years, deep learning has led to significant improvements in the performance of lip reading systems by enabling them to learn more representative and discriminative features directly from raw video frames~\cite{fernandez2018survey,prajwal2022sub,koumparoulis2022accurate}.
Initial breakthroughs in lip reading focused on word-level recognition~\cite{chung2017lip,stafylakis2017combining}. As the field progressed, ASR models based on LSTM sequence-to-sequence models~\cite{son2017lip} and CTC methods~\cite{shillingford2018large,assael2016lipnet} were developed for sentence-level recognition~\cite{zhao2020hearing,}. 
Recently, most related works have used either the Transformer~\cite {vaswani2017attention,kim2022distinguishing,shi2022robust} or Conformer~\cite{gulati2020conformer,ma2021lira} hybrid architectures. In a recent paper, Ma et al.~\cite{ma2023auto} leveraged a pre-trained ASR model for transcription and employed Conformer architecture to achieve 0.9$\%$WER on LRS3~\cite{afouras2018lrs3}.

Most relevant to our method is \cite{wang2023seeing}, which proposes using pre-trained lip reading experts to supervise training and using contrastive learning to enhance lip sync. While they use interpolation to supervise some of the frame data, this method cannot achieve frame-by-frame coherent supervision, and constraints on a small number of frames are not strong enough to significantly improve accuracy. In contrast, our approach utilizes a self-supervised frame-by-frame constraint, which provides a strong constraint and effectively improves the accuracy of lip movements.
\section{Method}
\subsection{Overview}
In this section, we describe methods utilized in our proposed SelfTalk model in detail. The key original idea of this model is to construct a cross-supervised training diagram with three components: facial animator, speech recognizer, and lip-reading interpreter. The commutativity of the diagram means the equivalent results from audio-recognizing and lip-reading, which ensures the facial reconstruction in quality from speech audio. The whole pipeline of our framework is revealed in Fig. \ref{fig:2}.

Technically, the facial animator regresses sequences of parametric expression vectors from speech segments, while the speech recognizer converts audio features the textual representations. On the other hand, the lip-reading interpreter maps the lip shape sequence to textual representations. The training loss constrains the distance between the output of lip-reading interpreter and its correspondence from the speech recognizer.

The SelfTalk framework can be formulated as follows: Let $\boldsymbol{Y}_{1: T}=\left(\boldsymbol{y}_{1}, \ldots, \boldsymbol{y}_{T}\right)$, $\forall \boldsymbol{y}_{t} \in \mathbb{R}^{V \times 3}$ be the sequence of vertices bias over a template face mesh $\boldsymbol{m}$ with $V$,  which is the face animation expected to be generated. In addition, the framework uses a sequence of speech snippets, $\boldsymbol{A}_{1: T}=\left(\boldsymbol{a}_{1}, \ldots, \boldsymbol{a}_{T}\right)$, where $ \boldsymbol{a}_{t} \in \mathbb{R}^{D}$ is a $D$-dimensional feature that aids in audio segmentation. The speech recognizer generates the sequence of sentence information expected, whereas the lip-reading interpreter creates text distribution, which is given by $\boldsymbol{S}_{1:T}=\left(\boldsymbol{s}_{1}, \ldots, \boldsymbol{s}_{T}\right)$ and $\boldsymbol{E}_{1:T}=\left(\boldsymbol{e}_{1}, \ldots, \boldsymbol{e}_{T}\right)$, respectively. Here, $\boldsymbol{s}_{t} \in \mathbb{R}^{U}$ and $\boldsymbol{e}_{t} \in \mathbb{R}^{U}$, where $U$ denotes the dimension of the output vocabulary size, which is 33.


The facial animator of SelfTalk regresses $\boldsymbol{\hat{Y}}_{1: T}\simeq\boldsymbol{Y}_{1: T}$ from $\boldsymbol{A}_{1: T}$ , which drives the 3D facial animation by speech audio. The equation that represents this process is:
\begin{equation}\label{1}
\boldsymbol{\hat{y}}_{t}=\mathrm{Facial~Animator}_{\theta}(\boldsymbol{a}_t),
\end{equation}
where $\theta$ represents the model parameters of the facial animator.

The speech recognizer learns the mapping from $\boldsymbol{A}_{1: T}$ to $\boldsymbol{S}_{1: T}$ and provides pseudo-ground-truth labels to train the commutative diagram. The formula is expressed as:
\begin{equation}\label{1}
\boldsymbol{s}_{t}=\mathrm{Speech~Recognizer}_{\phi}(\boldsymbol{a}_t),
\end{equation}
where the parameter $\phi$ represents the parameters of the speech recognizer.

The lip-reading interpreter comprehends the text content from the facial movements $\boldsymbol{\hat{Y}}_{1:T}$ and corresponds it with the output of the speech recognizer $\boldsymbol{E}_{1:T}$. Formally, we define: 
\begin{equation}\label{1}
\boldsymbol{e}_{t}=\mathrm{LipReading~Interpreter}_{\psi}(\boldsymbol{\hat{y}}_t),
\end{equation}
where $\psi$ represents the model parameters of the lip-reading interpreter. In the following parts of this section, we will provide detailed descriptions of each SelfTalk framework component.

\subsection{Facial Animator}
The facial animator module generates realistic and comprehensible facial expressions from audio input by utilizing self-supervised cross-modal constraints. Our design choice stems from the observation that existing speech-driven 3D face animation approaches frequently fail to generate visually comprehensible lip movements that can impact the animation's overall quality. Therefore, we use self-supervised constraints that capture the complex relationships among text, speech, and lip shape to generate more realistic and accurate facial expressions.

To attain this objective, the facial animator module comprises multiple sub-modules that collaborate. Firstly, we utilize the state-of-the-art self-supervised pre-trained speech model, wav2vec 2.0~\cite{baevski2020wav2vec}, for audio feature extraction. During the fine-tuning phase of the pre-trained model, we make the TCN~\cite{lea2016temporal} layer stationary since it is trained on an amount of audio data. Other layers can contribute to the model training and continually change to extract the desired features. This guarantees that the extracted audio features contain vast information about the spoken words. The formula is defined as follows:
\begin{equation}\label{2}
\boldsymbol{x}_{e1,t} = \mathrm{Audio~Encoder_1}(\boldsymbol{a}_t; \theta_{e}),
\end{equation}
where $\boldsymbol{x}_{e1,t}$ represents the audio features extracted by the first audio encoder, $\theta_{e}$ are the parameters of this audio encoder.

The extracted audio features are then fed into the mesh decoder.
The mesh decoder's function is to learn how to map audio embeddings to 3D mesh deformation in order to produce facial expressions aligned with the spoken words. This process is crucial to produce visually comprehensible lip movements that sync up with the audio input. We design the decoder using a transformer-based~\cite{vaswani2017attention} structure, a technique frequently utilized in natural language processing that has also proved effective in computer vision tasks.

The mesh decoder comprises several layers of masked self-attention and feed-forward neural networks. The masked self-attention mechanism enables the decoder to focus on relevant parts of the input sequence to produce suitable outputs. Specifically, the self-attention layer computes a weighted sum of the input embeddings based on their relevance to each other. Then it applies a feed-forward neural network to the resulting context vectors to generate the final output. The equations for this mechanism are as follows:
\begin{equation}\label{3}
\boldsymbol{h}^{(i)} = \mathrm{Multi~Head}(Q^{(i)}, K^{(i)}, V^{(i)}),
\end{equation}
\begin{equation}\label{4}
\mathrm{FFN}(\boldsymbol{h}^{(i)}) = \mathrm{ReLU}(\boldsymbol{h}^{(i)}\boldsymbol{W}_{f,1} + \boldsymbol{b}_f)\boldsymbol{W}_{f,2} + \boldsymbol{b}_f,
\end{equation}

where $Q$, $K$, and $V$ represent the query, key, and value matrices; $i$ denotes the index of the layer; $\boldsymbol{W}_{f,1}$, $\boldsymbol{W}_{f,2}$ are learnable parameters of the feed-forward neural network, and $\boldsymbol{b}_f$ is the bias term.

Finally, the mesh decoder takes the audio features as input and outputs a 3D mesh deformation, which can be represented as:
\begin{equation}\label{2}
\boldsymbol{\hat{y}} = \mathrm{Mesh~Decoder}(\boldsymbol{x}_{e1,t}; \theta_d),
\end{equation}

where $\theta_d$ are the parameters of the decoder, and $\boldsymbol{\hat{y}}$ indicates the produced 3D mesh deformation.

\subsection{Speech Recognizer}
The speech recognizer module is to extract informative latent features from the input audio signal and output text data, which is subsequently used to create self-supervised text feature constraints. We leverage the wav2vec 2.0~\cite{baevski2020wav2vec} pre-training model to pursue this goal. It uses a novel self-supervised learning technique similar to masked language modeling. In this approach, audio signals are transformed into waveform embeddings and then masked in certain areas to enable the model to predict the missing audio segments. This technique effectively extracts rich audio features that can be translated accurately into text.\par
The speech features of the input audio signal are extracted by applying a series of conversions and neural networks. First, the audio sampling rate is converted to 16 kHz. Then, the audio passes through seven feature encoders. Each encoder consists of temporal convolutions, which contain 512 channels. These convolutions have different strides and kernel widths. As a result, the encoder output frequency is 49 Hz with a stride of 20ms between each sample. This results in a receptive field of 400 input samples or 25ms of audio. The extracted features are fed into 24 transformer~\cite{vaswani2017attention} blocks, each with a 1,024 model dimension, 4,096 inner dimension, and 16 attention heads. The audio encoder of the speech recognizer can be defined as:
\begin{equation}\label{2}
\boldsymbol{x}_{e2,t} = \mathrm{Audio~Encoder_2}(\boldsymbol{a}_t; \phi_e),
\end{equation}
where $\boldsymbol{x}_{e2,t}$ represents the audio features extracted by the second audio encoder, $\phi_e$ are the parameters of this audio encoder.

After completing the audio feature extraction, we obtain latent features containing information about the input audio signal's content, emotion, and speaking style. The latent features are then passed into a pre-trained, fully connected language model decoder that maps 1024-dimensional features frame by frame into a 32-dimensional vocabulary consisting of 26 letters and six special characters such as "pad" and "star." We obtain the corresponding characters in the vocabulary by selecting the highest probability of the feature's output. Finally, a post-processing step is added to select words to remove duplicate letters and generate the final text. Formally, we define:

\begin{equation}\label{1}
\boldsymbol{s}_{t}=\mathrm{Text~Decoder_1}(\boldsymbol{x}_{e2,t}; \phi_d),
\end{equation}
where $\boldsymbol{s}_t$ represents the predicted transcription at time $t$, and the parameter $\phi_d$ represents the parameters of this text decoder. The speech recognizer module provides the network framework with latent and text features from a pre-trained model. This allows the model to self-supervise and train without needing ground-truth text input.
\subsection{Lip-Reading Interpreter}
The lip-reading interpreter module is critical in enabling our framework to implement the commutative training diagram. We obtain the mesh deformation output through the mesh decoder, then select corresponding lip vertices using lip indexes. We map lip vertex information to lip deformation features using a fully connected layer. 
For the lip encoding features to learn better feature expressions, we utilize a lip encoder module based on the Transformer~\cite{vaswani2017attention} framework. This module has six layers consisting of Transformer encoders that encode the lip features to extract its latent features. The latent features of the lip can be thought of as a point in the lip feature space that can represent different states of lip shape. We use the latent consistency loss function to constrain the latent features obtained from the lip encoding to be as similar as possible to the latent features extracted from the audio. The formula is defined as follows:
\begin{equation}\label{1}
\boldsymbol{l}_t=\mathrm{Lip~Encoder}(\boldsymbol{\hat{y}}; \psi_{e}),
\end{equation}
where $\boldsymbol{l}_t$ represents the lip features, and the parameter $\psi_{e}$ represents the parameters of the lip encoder. 
We create a text decoder module similar to the speech recognizer module, which can map 1024-dimensional latent features to a 32-dimensional vocabulary space. The text decoder is realized with fully connected layers,
\begin{equation}\label{1}
\boldsymbol{e}_{t}=\mathrm{Text~Decoder_2}(\boldsymbol{l}_t; \psi_{d}),
\end{equation}
where the parameter $\psi_{d}$ represents the parameters of this text decoder. 
To align the generated text features obtained from lip reading with the audio's original speech content, we compare the text features that the text decoder outputs with those of the speech recognizer module, using the connectionist temporal classification loss function, as detailed in Sec. \ref{sec3.5}.

We establish a commutative training diagram by combining the facial animator, speech recognizer, and lip-reading interpreter modules. Specifically, we generate text using the speech recognizer and facial animator modules to generate the corresponding facial mesh deformation. Then, the lip-reading interpreter module is trained using these inputs. After that, we apply the lip reading module to generate new text features from the mesh deformation, which we compare with the text features produced by the speech recognizer module to train the commutative diagram framework.

\subsection{Loss Function}
\label{sec3.5}
To train our SelfTalk framework, we employ a loss function that comprises four distinct components: reconstruction loss, velocity loss, latent consistency loss, and text consistency loss. The total loss function is a weighted sum of the four losses:

\begin{equation}\label{6}
L=\lambda_{1}L_{rec}+\lambda_{2}L_{vel}+\lambda_{3}L_{lat}+\lambda_{4}L_{ctc},
\end{equation}
where $\lambda_{1}$ = 1000.0, $\lambda_{2}$ = 1000.0, $\lambda_{3}$ = 0.001 and $\lambda_{4}$ = 0.0001 in all of our experiments. We provide a detailed explanation of each of these components below.\par
\noindent\textbf{Reconstruction Loss.}
The reconstruction loss measures the difference between the predicted lip movements and the ground-truth lip movements. Specifically, we use per-frame mean squared error (MSE) as the reconstruction loss:
\begin{equation}\label{7}
L_{rec} = \frac{1}{T} \sum_{t=1}^{T} \frac{1}{V} \sum_{v=1}^{V}\|\boldsymbol{y}_{t, v}-\boldsymbol{\hat{y}}_{t, v}\|^{2},
\end{equation}
where $T$ is the length of the sequence, $V$ is the number of vertices of the 3D face mesh.

\noindent\textbf{Velocity Loss.} To address the issue of jittery output frames when using only reconstruction loss, we introduce a velocity loss to encourage smooth and natural lip movements over time. The velocity loss can be expressed as:
\begin{equation}\label{8}
L_{vel} = \frac{1}{T} \sum_{t=1}^{T} \frac{1}{V} \sum_{v=1}^{V}\|\left(\boldsymbol{y}_{t, v}-\boldsymbol{y}_{t-1, v}\right) - \left(\hat{\boldsymbol{y}}_{t, v}-\hat{\boldsymbol{y}}_{t-1, v}\right)\|^{2},
\end{equation}
\noindent\textbf{Latent Consistency Loss.}
The latent consistency loss enables the model to learn more consistent feature representation from the training data, which measures the difference between the latent features extracted from the audio and the lip encoder using MSE. Additionally, this loss function aligns the learned audio and lip features, which helps with information exchange between audio and lip shape and constrains the formation of more comprehensible lip movements. The formula for calculating the latent consistency loss is:
\begin{equation}\label{9}
{L}_{lat} = \frac{1}{T} \sum_{t=1}^T \frac{1}{N}\sum_{i=1}^{N}\left \| \boldsymbol{x}_{e2,t,n} - \boldsymbol{l}_{t,n} \right \|_2^2,
\end{equation}
where $\boldsymbol{x}_{e2,t,n}$ represents the features extracted from the audio input $\boldsymbol{a}_{t, n}$, while $\boldsymbol{l}_{t,n}$ represents the features extracted from the lip encoder. $N$ denotes the number of samples.

\begin{figure*}[h]
  \centerline{\includegraphics[width=1.03\textwidth]{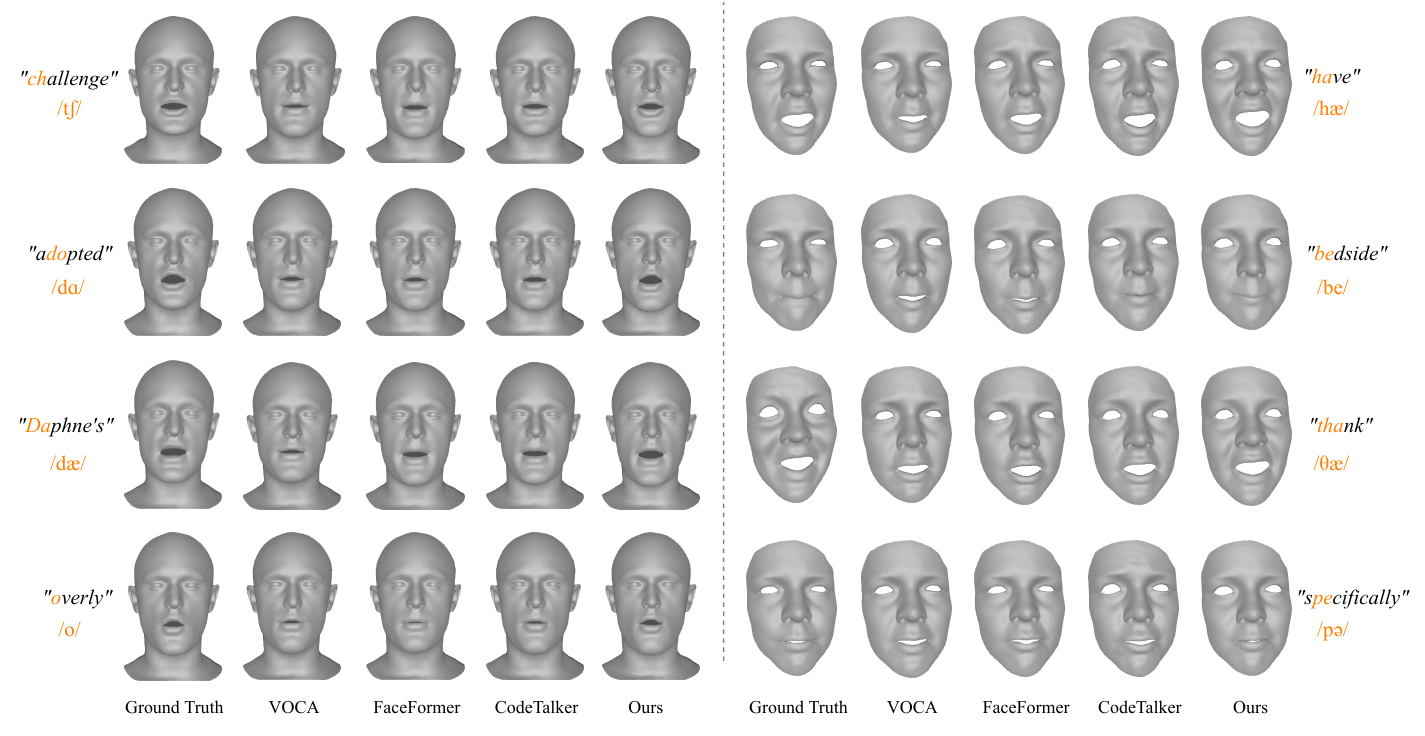}}
  \caption{Visual comparisons of facial movement by different methods on VOCA-Test (left) and BIWI-Test-B (right). We compared different syllables and found that our method produced more precise mouth movements. Specifically, our method showed a more noticeable upward curling movement of the upper lip for syllables that require opening the mouth, such as "/$\mathrm{d}$\textipa{A}/." For syllables that require closing the mouth, such as "/$\mathrm{be}$/," our method demonstrated more accurate closing movements and accurately learned the movement of the upper lip covering the lower lip.}
  \Description{SelfTalk}
  \label{fig:3}
\end{figure*}

\noindent\textbf{Text Consistency Loss.}
In order to ensure that the generated lip movements are coherent with the source audio and comprehensible for lip-reading purposes, we have introduced a text consistency loss calculation in which the discrepancy between the outputs of the lip-reading decoder and the original text is measured. To achieve this, we incorporated the Connectionist Temporal Classification (CTC) loss~\cite{graves2006connectionist} to train the lip-reading decoder. We used it to gauge the variation between the predicted character sequence and the ground truth text.

Let $\boldsymbol{E}_{1:T}$ represent the feature sequence extracted from the lip-reading module, and let  $\boldsymbol{S}_{1: T}=\left(\boldsymbol{s}_{1}, \ldots, \boldsymbol{s}_{t}\right)$ represent the pseudo ground truth transcription generated by the speech recognizer. For CTC, a projection layer is first used to map the lip-reading feature sequence to the output probabilities: 
\begin{equation}\label{10}
\boldsymbol{p}_t=\text{softmax}(\boldsymbol{W}^{ft}\boldsymbol{e}_t+\boldsymbol{b}^{ft}), 
\end{equation}
where $\boldsymbol{W}^{ft}\in\mathbb{R}^{d\times(U)}$, $\boldsymbol{b}^{ft}\in\mathbb{R}^{U}$, and $U$ is the output vocabulary size with a dimension of 33. The CTC loss can be defined as:
\begin{equation}\label{10}
L_{ctc}=-\log\sum_{\pi\in\mathcal{B}^{-1}(\boldsymbol{S})}p(\pi|\boldsymbol{e}_{1:T}),
\end{equation}
where $\mathcal{B}$ maps the alignment sequence $\pi$ to the transcription $\boldsymbol{S}$.
\subsection{Training Details}
The network is designed to receive audio and the corresponding facial mesh vertex template as inputs. In the pre-processing phase, the audio sampling rate is first converted to 16kHz, followed by the alignment of facial animation to match the audio. The VOCASET~\cite{cudeiro2019capture} is processed at 30fps, while BIWI~\cite{fanelli20103} remains at 25fps.

For model optimization during the training phase, we have employed the Adam optimizer~\cite{kingma2014adam} and have set the learning rate and batch size to $1e-4$ and 1, respectively. The entire network was trained on an NVIDIA RTX 3090, and it took around 2 hours (100 epochs) to complete the procedure.
\section{Experiments}

\subsection{Datasets}
We conducted training and testing utilizing two publicly available 3D datasets – BIWI~\cite{fanelli20103} and VOCASET~\cite{cudeiro2019capture}. Both datasets include audio-3D scan pairs showcasing English speech pronunciation. VOCASET features 255 unique sentences, some of which are shared across speakers. In contrast, BIWI includes 40 unique sentences shared across all speakers, making the task more challenging due to the less phonemic information provided.

\noindent\textbf{VOCASET Dataset.} VOCASET contains 480 facial motion sequences from 12 subjects, captured at 60fps for approximately 4 seconds per sequence. Each 3D face mesh is registered to the FLAME~\cite{li2017learning} topology with 5023 vertices. We adopted the same training (VOCA-Train), validation (VOCA-Val), and testing (VOCA-Test) splits as FaceFormer~\cite{fan2022faceformer} and CodeTalker~\cite{xing2023codetalker} for fair comparisons.

\noindent\textbf{BIWI Dataset.} BIWI is a 3D audio-visual corpus that displays emotional speech and facial expressions in dynamic 3D face geometry. It comprises two parts, one with emotions and the other devoid of them, with 40 sentences uttered by 14 subjects – six males and eight females. Each recording is repeated twice in neutral or emotional situations, capturing a dynamic 3D face at a 25fps rate. The registered topology exhibits 23370 vertices, with an average sequence length of around 4.67 seconds. We followed the data split from FaceFormer~\cite{fan2022faceformer} and CodeTalker~\cite{xing2023codetalker} and used the emotional subset only. Specifically, the training set (BIWI-Train) contains 192 sentences (actually 190, as two facial data were lost officially), while the validation set (BIWI-Val) contains 24 sentences. There are two testing sets, where BIWI-Test-A includes 24 sentences spoken by 6 participants seen during the training process and is suitable for both quantitative and qualitative evaluation. Contrarily, BIWI-TEST-B contains 32 sentences spoken by eight unseen participants and is more suitable for qualitative evaluation.

\noindent\textbf{Baseline Implementations.} We conducted experiments on BIWI and VOCASET datasets to compare the performance of SelfTalk with state-of-the-art methods that include VOCA~\cite{cudeiro2019capture}, MeshTalk~\cite{richard2021meshtalk}, FaceFormer~\cite{fan2022faceformer}, and CodeTalker~\cite{xing2023codetalker}. We trained and tested VOCA on BIWI using its official codebase and directly tested the released model trained on VOCASET. For MeshTalk, we used its official implementation to train and test on both datasets. For FaceFormer and CodeTalker, we used the provided pre-trained weights for testing.

\subsection{Quantitative Evaluation}
To measure lip synchronization, we calculated the lip vertex error (LVE) as used in MeshTalk~\cite{richard2021meshtalk} and FaceFormer~\cite{fan2022faceformer}, which is computed as the average $\ell_{2}$ error of the lips in the test set. However, LVE alone may not fully reflect the lip readability of the generated facial animation. Therefore, we propose a new evaluation metric called lip readability percentage (LRP), which can reflect the lip readability of the generated facial animation.
The lip readability percentage is defined as follows:
\begin{equation}\label{11}
\text{LRP} = \frac{1}{T} \sum_{t=1}^T \frac{1}{V} \sum_{i=1}^{V} [d_{t,v} < \mu] \times 100\%,
\end{equation}
where $d_{t, v}$ is the Euclidean distance between the $v^{th}$ lip vertex of the $t^{th}$ frame in the predicted lip sequence and the corresponding vertex in the ground truth lip sequence, $V_t$ is the number of lip vertices in the $t^{th}$ frame, and $\mu$ is the Euclidean distance threshold for determining whether a predicted lip shape is comprehensible. Specifically, if $d_{t, v} < \mu$, the predicted lip shape at time step $t$ is considered comprehensible, and otherwise, it is considered unreadable. This metric calculates the percentage of comprehensible lip shapes over the total number of lip shapes, reflecting the lip readability of the generated facial animation.

We follow the CodeTalker~\cite{xing2023codetalker} principles and utilize the Upper-face dynamics deviation metric. Facial expressions during speech are loosely correlated with the content and style of verbal communication, and as such, they vary significantly among individuals. The facial dynamics deviation (FDD) technique measures the dynamic changes in the face during a sequence of motions and compares them to ground truth.

We compared VOCA~\cite{cudeiro2019capture}, MeshTalk~\cite{richard2021meshtalk}, FaceFormer~\cite{fan2022faceformer}, \\CodeTalker~\cite{xing2023codetalker}, and our proposed method (SelfTalk) on BIWI and VOCASET datasets. We calculated the lip vertex error (LVE), the lip readability percentage (LRP), and the facial dynamics deviation (FDD) for all sequences in the BIWI-Test-A and VOCA-Test datasets.

According to Table \ref{tab:1} and \ref{tab:2}, our proposed SelfTalk method achieved lower errors and higher lip readability levels than the other methods evaluated. This suggests that SelfTalk captures the liveliness of the whole face while improving the accuracy and comprehensibility of lip movement. In particular, our lip vertex error on the VOCA-Test dataset is 18$\%$ lower than the newly proposed CodeTalker~\cite{xing2023codetalker} method, which strongly proves the effectiveness of our self-supervised commutative training diagram.

\begin{table}[]
\caption{Quantitative evaluation results on VOCA-Test.}
\label{tab:1}
\begin{tabular}{@{}lccc@{}}
\toprule
Methods         & \begin{tabular}[c]{@{}c@{}}LVE$\downarrow$\\ ($\times 10^{-5}$ mm)\end{tabular}               & \begin{tabular}[c]{@{}c@{}}FDD$\downarrow$\\ ($\times 10^{-7}$ mm)\end{tabular} & LRP$\uparrow$\\ \midrule
VOCA~\cite{cudeiro2019capture}            & 4.9245                                                                       & 4.8447              & 72.67\%                                                 \\
MeshTalk~\cite{richard2021meshtalk}        & 4.5441                                                                        & 5.2062             & 79.64\%                                                 \\
FaceFormer~\cite{fan2022faceformer}      & 4.1090                                                                         & 4.6675            & 88.90\%                                                   \\
CodeTalker~\cite{xing2023codetalker}      & 3.9445                                                                        & 4.5422             & 86.30\%                                                 \\
SelfTalk (Ours) & \textbf{3.2238}                                                      & \textbf{4.0912}             & \textbf{91.37\%}                                        \\ \bottomrule
\end{tabular}
\end{table}

\begin{table}[]
\caption{Quantitative evaluation results on BIWI-Test-A.}
\label{tab:2}
\begin{tabular}{@{}lccc@{}}
\toprule
Methods                & \begin{tabular}[c]{@{}c@{}}LVE$\downarrow$\\ ($\times 10^{-4}$ mm)\end{tabular}            & \begin{tabular}[c]{@{}c@{}}FDD$\downarrow$\\ ($\times 10^{-5}$ mm)\end{tabular}  & LRP$\uparrow$       \\ \midrule
VOCA~\cite{cudeiro2019capture}            & 6.5563                                                                    & 8.1816                                   & 73.83\%                               \\
MeshTalk~\cite{richard2021meshtalk}        & 5.9181                                                              & 5.1025                                              & 80.97\%                          \\
FaceFormer~\cite{fan2022faceformer}      & 5.3077                                                                 & 4.6408                                        & 83.15\%                             \\
CodeTalker~\cite{xing2023codetalker}      & 4.7914                                                                    & 4.1170                                         & 84.62\%                         \\
SelfTalk (Ours) & \textbf{4.2485}                                                    & \textbf{3.5761}                     & \textbf{88.31\%}                                  \\ \bottomrule
\end{tabular}
\end{table}

\subsection{Qualitative Evaluation}
While metrics are essential for evaluating 3D talking faces, visualizing the prediction results is necessary for a more comprehensive understanding of the model's performance. Thus, we introduced qualitative evaluation to compare our model with others. In this comparison process, for a fair evaluation, we assigned the same speaking style as the conditional input for VOCA~\cite{cudeiro2019capture}, FaceFormer~\cite{fan2022faceformer}, and CodeTalker~\cite{xing2023codetalker}. We recommend that readers watch our supplementary video for prediction performance evaluation of multiple methods, including our proposed technique and other previous methods~\cite{cudeiro2019capture,fan2022faceformer,xing2023codetalker}, as well as for the ground truth. To conduct the evaluation, audio sequences from BIWI and VOCASET test sets were utilized, and audio clips were extracted from other methods' supplementary videos for comparison. Furthermore, we extracted some clips from YouTube to test our model's ability to generalize to in-the-wild audio and to generalize to multiple languages. The demonstration video showcases that SelfTalk generates coherent and realistic semantic movements, offering high lip movement comprehensibility. Even in cases of rapid lip movement, our framework can accurately generate significant lip shape changes.\par
\begin{figure}[t]
  \centerline{\includegraphics[width=0.45\textwidth]{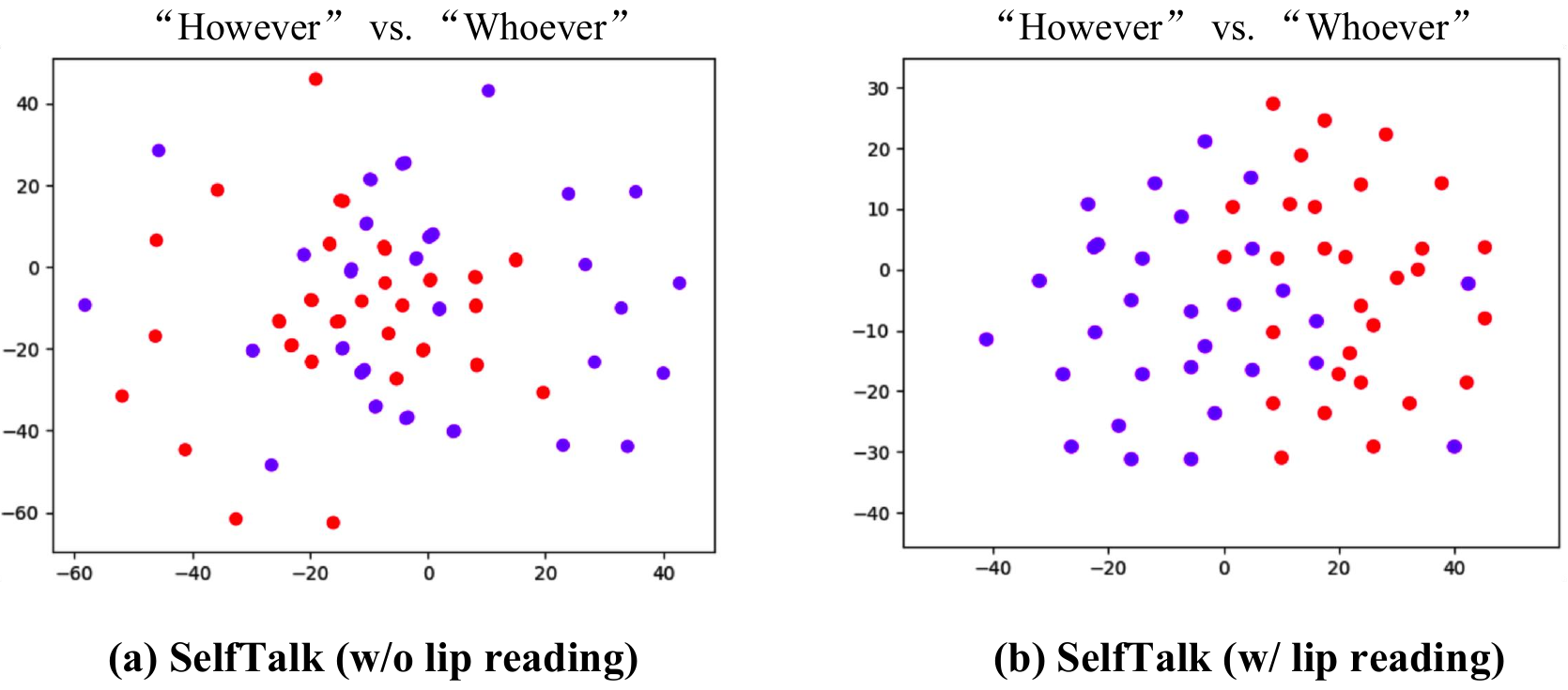}}
  \caption{The t-SNE visualization of the lip feature corresponds to "However" (red) and "Whoever" (purple). The left image displays a mixture of different features since no lip-reading module was added, while the right image, containing the lip-reading module, distinguishes the different word features.}
  \Description{The t-SNE visualization.}
  \label{fig:4}
\vspace{-5pt}
\end{figure}

 
In addition, we compared the effects of different models in Fig. \ref{fig:3}, where facial animation clips were captured of people speaking specific phonemes. Compared with other methods, our model resembles the ground truth and can generate more accurate lip movements. For example, our method demonstrates a more pronounced puckering movement when producing the syllable "/\textteshlig/." When producing the syllable "/$\mathrm{h}$\ae/," our method has the more pronounced upward movement of the upper lip.
Additionally, we employed t-SNE~\cite{van2008visualizing} to visualize the lip features both with and without lip-reading constraint, as illustrated in Fig. \ref{fig:4}. This visualization makes it more evident that with the introduction of lip-reading constraints, different feature sets of words are distinguished, further proving that lip-reading constraints are effective in guiding lip movement.

\begin{table}[]
\caption{User study results on BIWI-Test-B and VOCA-Test.}
\label{tab:3}
\begin{center}
\resizebox{\linewidth}{!}{
\begin{tabular}{lcccc}
\toprule
\multirow{2}{*}{Method}  & \multicolumn{2}{c}{{ BIWI-Test-B}}   & \multicolumn{2}{c}{{ VOCA-Test}}    \\
                          & competitor      & ours           & competitor      & ours  \\ \hline
\textbf{Ours vs. MeshTalk}   &                 &                 \\
Lip Sync                     & 24.4\%          & \textbf{75.6\%}   & 21.6\%          & \textbf{78.4\%}\\
Realism                      & 25.9\%          & \textbf{74.1\%}   & 24.4\%          & \textbf{75.6\%}\\ \midrule
\textbf{Ours vs. FaceFormer} &                 &                 \\
Lip Sync                     & 34.4\%          & \textbf{65.6\%}   & 30.5\%          & \textbf{69.5\%}\\
Realism                      & 35.8\%          & \textbf{64.2\%}   & 33.3\%          & \textbf{66.7\%}\\ \midrule
\textbf{Ours vs. CodeTalker}   &                 &                 \\
Lip Sync                     & 37.9\%          & \textbf{62.1\%}   & 32.8\%          & \textbf{67.2\%}\\
Realism                      & 40.6\%          & \textbf{59.4\%}   & 34.1\%          & \textbf{65.9\%}\\ \bottomrule
\end{tabular}
}
\end{center}
\end{table}
\subsection{User Study}
A user study is a reliable evaluation method in 3D talking faces. To compare our method with MeshTalk~\cite{richard2021meshtalk}, FaceFormer~\cite{fan2022faceformer}, and CodeTalker~\cite{xing2023codetalker}, we conducted a user study like~\cite{fan2022faceformer} to evaluate two metrics, perceptual lip synchronization, and facial realism. Our study included a comparison with the side-by-side presentation, and users chose the more realistic facial animation based on their preference. We calculated the ratio of user choices for satisfaction evaluation.
Table \ref{tab:3} shows that our method exhibited better perceptual lip synchronization and facial realism. For instance, 67.2$\%$ of users preferred our lip synchronization method on VOCA-Test compared to CodeTalker. The lip-reading module used in our method acted as a strict constraint that facilitated better semantic comprehension and more natural realism. 
\begin{table}[]
\caption{Ablation study for our components on BIWI-Test-A.}
\label{tab:freq}
\begin{tabular}{@{}lccc@{}}
\toprule
                              & LVE$\downarrow$                         & FDD$\downarrow$        & LRP$\uparrow$       \\ \midrule
Ours                          & \textbf{4.2485}  & \textbf{3.5761} & \textbf{88.31\%}\\\midrule
w/o $L_{vel}$ Loss            & 4.6655                    & 4.0681 & 84.15\%         \\
w/o $L_{lat}$ Loss            & 4.6846                    & 3.6114 & 86.88\%         \\
w/o $L_{text}$ Loss           & 4.9168                   & 3.7993  & 85.87\%        \\
w/o Commutative Training Diagram        & 5.3772              & 4.0378  & 83.08\%      \\ \bottomrule
\end{tabular}
\end{table}

\subsection{Ablation Study}
In this section, we conduct an ablation study to investigate the impact of various elements in our proposed SelfTalk framework on the quality of generated 3D talking faces. Specifically, we study the impact of removing the velocity loss, latent consistency loss, text consistency loss, commutative training diagram.

\noindent\textbf{Impact of the Velocity Loss.} We investigate the effect of removing velocity loss, which encourages predicted facial animations to match the ground truth's movement velocity closely. Removing the velocity loss significantly deteriorated all indicators, particularly FDD, which increased by approximately 12$\%$. This indicates that velocity loss enables more precise dynamic changes in the upper part of the face. Meanwhile, the removal of the velocity loss resulted in an inter-frame jitter observed in the generated faces. As such, the findings suggest that the velocity loss can introduce prior facial motion based on ground truth labels, which can produce more coherent facial animations.

\noindent\textbf{Impact of the Latent and Text Consistency Loss.} We conducted a detailed study of the impact of removing the latent and text consistency losses by removing them from our whole model. The latent consistency loss ensures that the lip embeddings generated by our network are consistent over time and aligned with the original audio, while the text consistency loss ensures that the lip movements match the transcribed text. Ignoring these losses resulted in a substantial decrease in the overall visual quality of the generated faces and the comprehensibility of lip movements. Specifically, removing the text consistency loss led to an increase of 13.6$\%$ in LVE and caused the lip-reader module to fail to output any text.

\noindent\textbf{Impact of the Commutative Training Diagram.} The commutative training diagram constitutes the core of our framework. When removing the commutative training diagram, we only train the facial animator module. It has been found that the error rate of the generated facial animation lip shape increases significantly by over 20$\%$ in LVE. This observation suggests that data-driven regression models may encounter domain gaps across various modalities, leading to pronounced ambiguity between audio and facial features. This renders it difficult for the model to learn precise facial expressions. The effectiveness of our proposed method in addressing these issues is demonstrated in this section.

\section{Conclusion}
In this paper, we proposed SelfTalk, a novel framework for 3D talking faces that addresses the limitations of traditional regression models in generating realistic and accurate lip movements with lip-reading comprehensibility. SelfTalk incorporates self-supervision in a cross-modal network system consisting of three modules: facial animator, speech recognizer, and lip-reading interpreter. Through its training process, SelfTalk establishes a commutative diagram that facilitates latent feature exchange between audio, text, and lip shape to create realistic and comprehensible facial animations.
Experimental results and user studies showed that SelfTalk outperforms current methods regarding state-of-the-art performance in 3D talking face. Moreover, SelfTalk reduces the dependence on labeled data, making multimedia applications more efficient and flexible.

\begin{acks}
This work was supported in part by National Key Research and Development Program of China under Grant No. 2020YFB2104101, National Natural Science Foundation of China (NSFC) under Grant Nos. 62172421, 62072459 and 71771131, as well as the Public Computing Cloud at Renmin University of China.
\end{acks}

\bibliographystyle{ACM-Reference-Format}
\bibliography{sample-base}










\end{document}